\documentclass[english]{article}
\usepackage[latin9]{inputenc}
\usepackage{babel}
\usepackage{wrapfig}
\usepackage{url}
\usepackage{amstext}
\usepackage{graphicx}
\usepackage[authoryear]{natbib}
\usepackage[unicode=true,
 bookmarks=true,bookmarksnumbered=false,bookmarksopen=false,
 breaklinks=false,pdfborder={0 0 1},backref=false,colorlinks=false]
 {hyperref}

\makeatletter
\usepackage[final]{neurips_2019}

\@ifundefined{showcaptionsetup}{}{%
 \PassOptionsToPackage{caption=false}{subfig}}
\usepackage{subfig}
\makeatother

\begin{document}

\title{On the Ineffectiveness of Variance Reduced Optimization for Deep
Learning}

\author{Aaron Defazio \& L\'eon Bottou \\
 Facebook AI Research New York}
\maketitle
\begin{abstract}
The application of stochastic variance reduction to optimization has
shown remarkable recent theoretical and practical success. The applicability
of these techniques to the hard non-convex optimization problems encountered
during training of modern deep neural networks is an open problem.
We show that naive application of the SVRG technique and related approaches
fail, and explore why.\vspace{-0.5em}
\end{abstract}

\section{Introduction\vspace{-0.5em}
}

\label{sec:introduction} Stochastic variance reduction (SVR) consists
of a collection of techniques for the minimization of finite-sum problems:
\[
{\textstyle f(w)=\frac{1}{n}\sum_{i=1}^{n}f_{i}(w),}
\]
 such as those encountered in empirical risk minimization, where each
$f_{i}$ is the loss on a single training data point. Principle techniques
include SVRG \citep{svrg}, SAGA \citep{adefazio-nips2014}, and their
variants. SVR methods use control variates to reduce the variance
of the traditional stochastic gradient descent (SGD) estimate $f_{i}^{\prime}(w)$
of the full gradient $f^{\prime}(w)$. Control variates are a classical
technique for reducing the variance of a stochastic quantity without
introducing bias. Say we have some random variable $X$. Although
we could use $X$ as an estimate of $E[X]=\bar{X}$, we can often
do better through the use of a control variate $Y$. If $Y$ is a
random variable correlated with $X$ (i.e. $\text{Cov}[X,Y]>0$),
then we can estimate $\bar{X}$ with the quantity
\[
{\textstyle Z=X-Y+E[Y].}
\]
This estimate is unbiased since $-Y$ cancels with $E[Y]$ when taking
expectations, leaving $E[Z]=E[X]$. As long as $Var[Y]\leq2\text{Cov}[X,Y]$,
the variance of $Z$ is lower than that of $X$.

Remarkably, these methods are able to achieve linear convergence rates
for smooth strongly-convex optimization problems, a significant improvement
on the sub-linear rate of SGD. SVR methods are part of a larger class
of methods that explicitly exploit finite-sum structures, either by
dual (\citealp[SDCA,][]{SDCA}; \citealp[MISO,][]{miso2}; \citealp[Finito,][]{adefazio-icml2014})
or primal (\citealp[SAG,][]{SAG}) approaches.

Recent work has seen the fusion of acceleration with variance reduction
(\citet{accel-sdca,catalyst,adefazio-nips2016,katyusha}), and the
extension of SVR approaches to general non-convex \citep{pmlr-v48-allen-zhua16,reddi-nonconvex-svrg}
as well as saddle point problems \citep{bach-saddle}. 

In this work we study the behavior of variance reduction methods on
a prototypical non-convex problem in machine learning: A deep convolutional
neural network designed for image classification. We discuss in Section
\ref{sec:complications} how standard training and modeling techniques
significantly complicate the application of variance reduction methods
in practice, and how to overcome some of these issues. In Sections
\ref{sec:measuring} \& \ref{sec:var-red-fails} we study empirically
the amount of variance reduction seen in practice on modern CNN architectures,
and we quantify the properties of the network that affect the amount
of variance reduction. In Sections \ref{sec:streaming-svrg} \& \ref{sec:convergence-rate-comparisons}
we show that streaming variants of SVRG do not improve over regular
SVRG despite their theoretical ability to handle data augmentation.
Code to reproduce the experiments performed is provided on the first
author's website.

\section*{Standard SVR approach\vspace{-0.5em}
}

The SVRG method is the simplest of the variance reduction approaches
to apply for large-scale problems, so we will focus our initial discussion
on it. In SVRG, training epochs are interlaced with snapshot points
where a full gradient evaluation is performed. The iterate at the
snapshot point $\tilde{w}$ is stored, along with the full gradient
$f^{\prime}(\tilde{w})$. Snapshots can occur at any interval, although
once per epoch is the most common frequency used in practice. The
SGD step $w_{k+1}=w_{k}-\gamma f_{i}^{\prime}(w_{k}),$ using the
randomly sampled data-point loss $f_{i}$ with step size $\gamma$,
is augmented with the snapshot gradient using the control variate
technique to form the SVRG step:
\begin{equation}
w_{k+1}=w_{k}-\gamma\left[f_{i}^{\prime}(w_{k})-f_{i}^{\prime}(\tilde{w})+f^{\prime}(\tilde{w})\right].\label{eq:svrg_step}
\end{equation}

The single-data point gradient $f_{i}^{\prime}(\tilde{w})$ may be
stored during the snapshot pass and retrieved, or recomputed when
needed. The preference for recomputation or storage depends a lot
on the computer architecture and its bottlenecks, although recomputation
is typically the most practical approach. 

Notice that following the control variate approach, the expected step,
conditioning on $w_{k}$, is just a gradient step. So like SGD, it
is an unbiased step. Unbiasedness is not necessary for the fast rates
obtainable by SVR methods, both SAG \citep{SAG} and Point-SAGA \citep{adefazio-nips2016}
use biased steps, however biased methods are harder to analyze. Note
also that successive step directions are highly correlated, as the
$f^{\prime}(\tilde{w})$ term appears in every consecutive step between
snapshots. This kind of step correlation is also seen in momentum
methods, and is considered a contributing factor to their effectiveness
\citep{insuff_momentum}. 

\vspace{-0.5em}

\section{Complications in practice\vspace{-0.5em}
}

\label{sec:complications}

Modern approaches to training deep neural networks deviate significantly
from the assumptions that SVR methods are traditionally analyzed under.
In this section we discuss the major ways in which practice deviates
from theory and how to mitigate any complications that arise.

\subsection*{Data augmentation\vspace{-0.5em}
}

\begin{wrapfigure}{r}{0.5\columnwidth}%
\includegraphics[width=0.48\columnwidth]{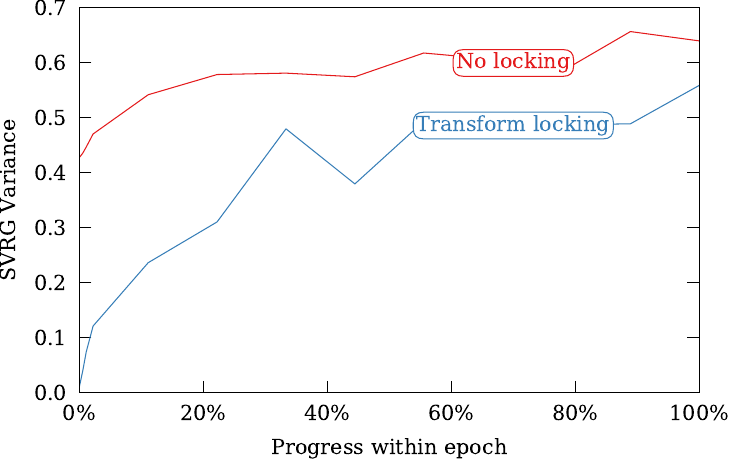}\caption{\label{fig:variance-locking}Variance within epoch two during LeNet
training on CIFAR10.\vspace{-0.5em}
}
\end{wrapfigure}%
In order to achieve state-of-the-art results in most domains, data
augmentation is essential. The standard approach is to form a class
of transform functions $\mathcal{T}$; for an image domain typical
transforms include cropping, rotation, flipping and compositions thereof.
Before the gradient calculation for a data-point $x_{i}$, a transform
$T_{i}$ is sampled and the gradient is evaluated on its image $T_{i}(x_{i})$.

When applying standard SVRG using gradient recomputation, the use
of random transforms can destroy the prospects of any variance reduction
if different transforms are used for a data-point during the snapshot
pass compared to the following steps. Using a different transform
is unfortunately the most natural implementation when using standard
libraries (PyTorch\footnote{\url{http://pytorch.org/}}; TensorFlow,
\citet{tensorflow2015-whitepaper}), as the transform is applied automatically
as part of the data-pipeline. We propose the use of transform \emph{locking},
where the transform used during the snapshot pass is cached and reused
during the following epoch/s.

This performance difference is illustrated in Figure \ref{fig:variance-locking},
where the variance of the SVRG step is compared with and without transform
locking during a single epoch during training of a LeNet model. Data
augmentation consisted of random horizontal flips and random cropping
to 32x32, after padding by 4 pixels on each side (following standard
practice).

For SVRG with transform locking, the variance of the step is initially
zero at the very beginning of the epoch, increasing over the course
of the epoch. This is the behavior expected of SVRG on finite sum
problems. In contrast, without transform locking the variance is non-zero
at the beginning of the epoch, and uniformly worse.

The handling of data augmentation in finite-sum methods has been previously
considered for the MISO method \citep{online-miso-mairal}, which
is one of the family of gradient table methods (as with the storage
variant of SVRG). The stored gradients are updated with an exponential
moving average instead of overwriting, which averages over multiple
past transformed-data-point gradients. As we show in Section \ref{sec:var-red-fails},
stored gradients can quickly become too stale to provide useful information
when training large models.

\vspace{-0.5em}

\subsection*{Batch normalization\vspace{-0.5em}
}

Batch normalization \citep{batchnorm} is another technique that breaks
the finite-sum structure assumption. In batch normalization, mean
and variance statistics are calculated within a mini-batch, for the
activations of each layer (typically before application of a nonlinearity).
These statistics are used to normalize the activations. The finite
sum structure no longer applies since the loss on a datapoint $i$
depends on the statistics of the mini-batch it is sampled in.

The interaction of BN with SVRG depends on if storage or recomputation
of gradients is used. When recomputation is used naively, catastrophic
divergence occurs in standard frameworks. The problem is a subtle
interaction with the internal computation of running means and variances,
for use at test time.

In order to apply batch normalization at test time, where data may
not be mini-batched or may not have the same distribution as training
data, it is necessary to store mean and variance information at training
time for later use. The standard approach is to keep track of a exponential
moving average of the mean and variances computed at each training
step. For instance, PyTorch by default will update the moving average
$m_{EMA}$ using the mini-batch mean $m$ as:
\[
m_{EMA}=\frac{9}{10}m_{EMA}+\frac{1}{10}m.
\]

During test time, the network is switched to evaluation mode using
\texttt{model.eval()}, and the stored running mean and variances are
then used instead of the internal mini-batch statistics for normalization.
The complication with SVRG is that during training the gradient evaluations
occur both at the current iterate $x_{k}$ and the snapshot iterate
$\tilde{x}$. If the network is in train mode for both, the EMA will
average over activation statistics between two different points, resulting
in \textbf{poor results and divergence}. 

Switching the network to evaluation mode mid-step is the obvious solution,
however computing the gradient using the two different sets of normalizations
results in additional introduced variance. We recommend a BN \emph{reset
}approach, where the normalization statistics are temporarily stored
before the $\tilde{w}$ gradient evaluation, and the stored statistics
are used to undo the updated statistics by overwriting afterwards.
This avoids having to modify the batch normalization library code.
It is important to use train mode during the snapshot pass as well,
so that the mini-batch statistics match between the two evaluations.\vspace{-0.5em}

\subsection*{Dropout\vspace{-0.5em}
}

Dropout \citep{dropout} is another popular technique that affects
the finite-sum assumption. When dropout is in use, a random fraction,
usually 50\%, of the activations will be zero at each step. This is
extremely problematic when used in conjunction with variance reduction,
since the sparsity pattern will be different for the snapshot evaluation
of a datapoint compared to its evaluation during the epoch, resulting
in much lower correlation and hence lower variance reduction. 

The same dropout pattern can be used at both points as with the transform
locking approach proposed above. The seed used for each data-point's
sparsity pattern should be stored during the snapshot pass, and reused
during the following epoch when that data-point is encountered. Storing
the sparsity patterns directly is not practical as it will be many
times larger than memory even for simple models.

Residual connection architectures benefit very little from dropout
when batch-norm is used \citep{resnet,batchnorm}, and because of
this we don't use dropout in the experiments detailed in this work,
following standard practice. \vspace{-0.5em}

\subsection*{Iterate averaging\vspace{-0.5em}
}

Although it is common practice to use the last iterate of an epoch
as the snapshot point for the next epoch, standard SVRG theory requires
computing the snapshot at either an average iterate or a randomly
chosen iterate from the epoch instead. Averaging is also needed for
SGD when applied to non-convex problems. We tested both SVRG and SGD
using averaging of 100\%, 50\% or 10\% of the tail of each epoch as
the starting point of the next epoch. Using a 10\% tail average did
result in faster initial convergence for both methods before the first
step size reduction on the CIFAR10 test problem (detailed in the next
section). However, this did not lead to faster convergence after the
first step size reduction, and final test error was consistently worse
than without averaging. For this reason we did not use iterate averaging
in the experiments presented in this work.\vspace{-0.5em}

\section{Measuring variance reduction\vspace{-0.5em}
}

\label{sec:measuring}To illustrate the degree of variance reduction
achieved by SVRG on practical problems, we directly computed the variance
of the SVRG gradient estimate, comparing it to the variance of the
stochastic gradient used by SGD. To minimize noise the variance was
estimated using a pass over the full dataset, although some noise
remains due to the use of data augmentation. The transform locking
and batch norm reset techniques described above were used in order
to get the most favorable performance out of SVRG.

Ratios below one indicate that variance reduction is occurring, whereas
ratios around two indicate that the control variate is uncorrelated
with the stochastic gradient, leading to an \emph{increase} in variance.
For SVRG to be effective we need a ratio below $1/3$ to offset the
additional computational costs of the method. We plot the variance
ratio at multiple points within each epoch as it changes significantly
during each epoch. An initial step size of 0.1 was used, with 10-fold
decreases at 150 and 220 epochs. A batch size of 128 with momentum
0.9 and weight decay 0.0001 was used for all methods. Without-replacement
data sampling was used.

\begin{figure*}[t]
\hfill{}\subfloat[{\scriptsize{}LeNet}]{\includegraphics[width=0.48\columnwidth]{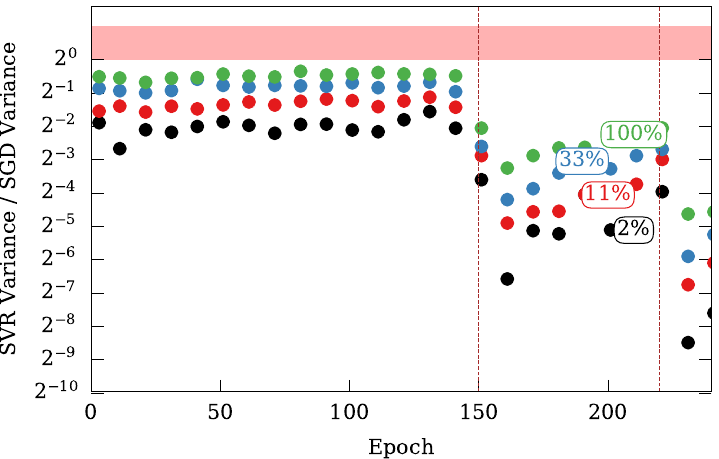}}\subfloat[{\scriptsize{}DenseNet-40-36}]{\includegraphics[width=0.48\columnwidth]{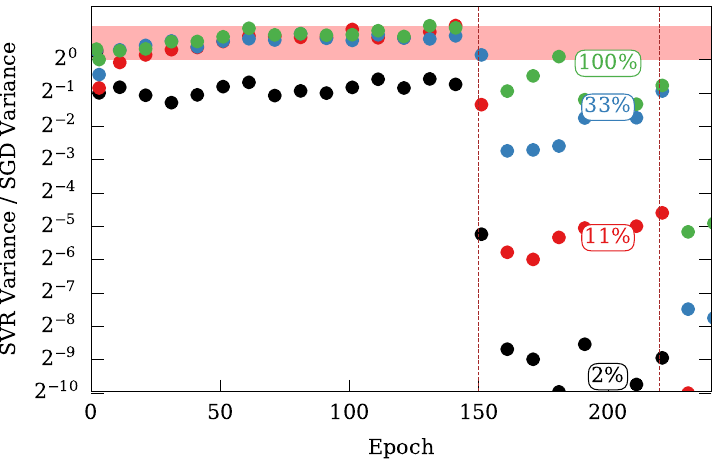}}\hfill{}\vspace{-1em}

\hfill{}\subfloat[{\scriptsize{}Small ResNet}]{\includegraphics[width=0.48\columnwidth]{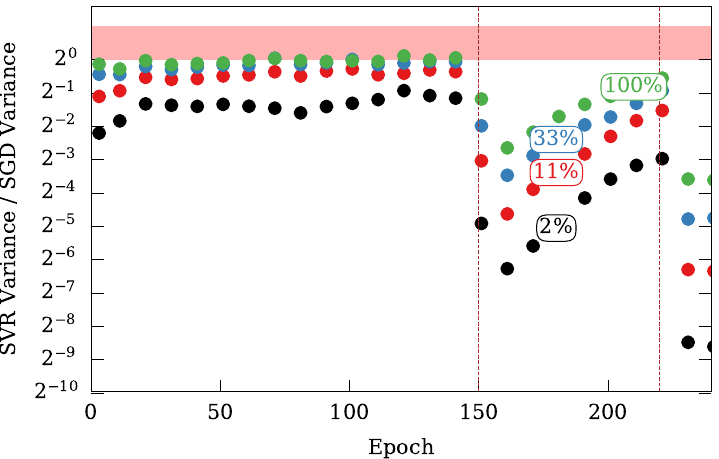}}\subfloat[{\scriptsize{}ResNet-110}]{\includegraphics[width=0.48\columnwidth]{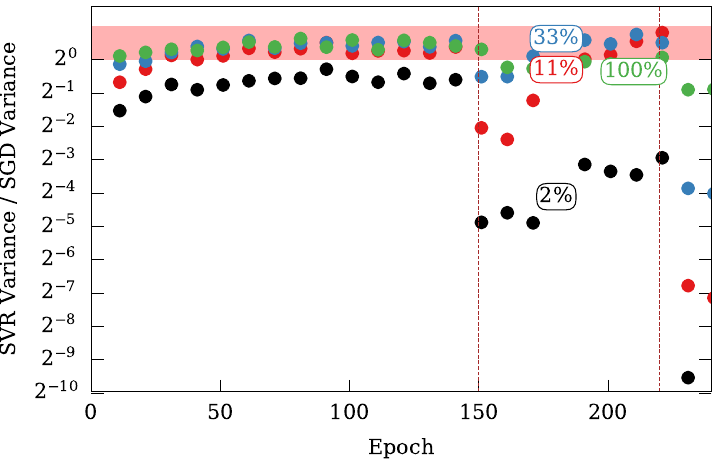}

}\hfill{}\vspace{-0.5em}

\caption{\label{fig:variance-ratios}The SVRG to SGD gradient variance ratio
during a run of SVRG. The shaded region indicates a variance \textit{increase},
where the SVRG variance is worse than the SGD baseline. Dotted lines
indicate when the step size was reduced. The variance ratio is shown
at different points within each epoch, so that the 2\% dots (for instance)
indicate the variance at 1,000 data-points into the 50,000 datapoints
consisting of the epoch. Multiple percentages within the same run
are shown at equally spaced epochs.\protect \\
SVRG fails to show a variance reduction for the majority of each epoch
when applied to modern high-capacity networks, whereas some variance
reduction is seem for smaller networks.}
\end{figure*}

To highlight differences introduced by model complexity, we compared
four models:
\begin{enumerate}
\item The classical LeNet-5 model \citep{lenet}, modified to use batch-norm
and ReLUs, with approximately 62 thousand parameters\footnote{Connections between max pooling layers and convolutions are complete,
as the symmetry breaking approach taken in the the original network
is not implemented in modern frameworks.}. 
\item A ResNet-18 model \citep{resnet}, scaled down to match the model
size of the LeNet model by halving the number of feature planes at
each layer. It has approximately 69 thousand parameters.
\item A ResNet-110 model with 1.7m parameters, as used by \citet{resnet}.
\item A wide DenseNet model \citep{densenet} with growth rate 36 and depth
40. It has approximately 1.5 million parameters and achieves below
5\% test error.
\end{enumerate}
Figure \ref{fig:variance-ratios} shows how this variance ratio depends
dramatically on the model used. For the LeNet model, the SVRG step
has consistently lower variance, from 4x to 2x depending on the position
within the epoch, during the initial phase of convergence. 

In contrast, the results for the DenseNet-40-36 model as well as the
ResNet-110 model show an \emph{increase} in variance, for the majority
of each epoch, up until the first step size reduction at epoch 150.
Indeed, even at only 2\% progress through each epoch, the variance
reduction is only a factor of 2, so computing the snapshot pass more
often than once an epoch can not help during the initial phase of
optimization. 

The small ResNet model sits between these two extremes, showing some
variance reduction mid-epoch at the early stages of optimization.
Compared to the LeNet model of similar size, the modern architecture
with its greater ability to fit the data also benefits less from the
use of SVRG.\vspace{-0.5em}

\section{Snapshot intervals\vspace{-0.5em}
}

The number of stochastic steps between snapshots has a significant
effect on the practical performance of SVRG. In the classical convex
theory the interval should be proportional to the condition number
\citep{svrg}, but in practice an interval of one epoch is commonly
used, and that is what we used in the experiment above. A careful
examination of our results from Figure \ref{fig:variance-ratios}
show that no adjustment to the snapshot interval can salvage the method.
The SVRG variance can be kept reasonable (i.e. below the SGD variance)
by reducing the duration between snapshots, however for the ResNet-110
and DenseNet models, even at 11\% into an epoch, the SVRG step variance
is already larger than that of SGD, at least during the crucial 10-150
epochs. If we were to perform snapshots at this frequency the wall-clock
cost of the SVRG method would go up by an order of magnitude compared
to SGD, while still under-performing on a per-epoch basis. 

Similarly, we can consider performing snapshots at less frequent intervals.
Our plots show that the variance of the SVRG gradient estimate will
be approximately 2x the variance of the SGD estimate on the harder
two problems in this case (during epochs 10-150), which certainly
will not result in faster convergence. This is because the correction
factor in Equation \ref{eq:svrg_step} becomes so out-of-date that
it becomes effectively uncorrelated with the stochastic gradient,
and since it's magnitude is comparable (the gradient norm decays relatively
slowly during optimization for these networks) adding it to the stochastic
gradient results in a doubling of the variance.

\subsection{Variance reduction and optimization speed}

For sufficiently well-behaved objective functions (such as smooth
\& strongly convex), we can expect that an increase of the learning
rate results in a increase of the converge rate, up until the learning
rate approaches a limit defined by the curvature ($\approx1/L$ for
L Lipschitz-smooth functions). This holds also in the stochastic case
for small learning rates, however there is an additional ceiling that
occurs as you increase the learning rate, where the variance of the
gradient estimate begins to slow convergence. Which ceiling comes
into effect first determines if a possible variance reduction (such
as from SVRG) can allow for larger learning rates and thus faster
convergence. Although clearly a simplified view of the non-differentiable
non-convex optimization problem we are considering, it still offers
some insight.

Empirically deep residual networks are known to be constrained by
the curvature for a few initial epochs, and afterwards are constrained
by the variance. For example, \citet{imagenet1h} show that decreasing
the variance by increasing the batch-size allows them to proportionally
increase the learning rate for variance reduction factors up to $30$
fold. This is strong evidence that a SVR technique that results in
significant variance reduction can potentially improve convergence
in practice. 

\section{Why variance reduction fails\vspace{-0.5em}
}

\label{sec:var-red-fails}Figure \ref{fig:variance-ratios} clearly
illustrates that for the DenseNet model, SVRG gives no actual variance
reduction for the majority of the optimization run. This also holds
for larger ResNet models (plot omitted). The variance of the SVRG
estimator is directly dependent on how similar the gradient is between
the snapshot point $\tilde{x}$ and the current iterate $x_{k}$.
Two phenomena may explain the differences seen here. If the $w_{k}$
iterate moves too quickly through the optimization landscape, the
snapshot point will be too out-of-date to provide meaningful variance
reduction. Alternatively, the gradient may just change more rapidly
in the larger model. 

\begin{figure*}
\begin{centering}
\includegraphics[width=0.48\columnwidth]{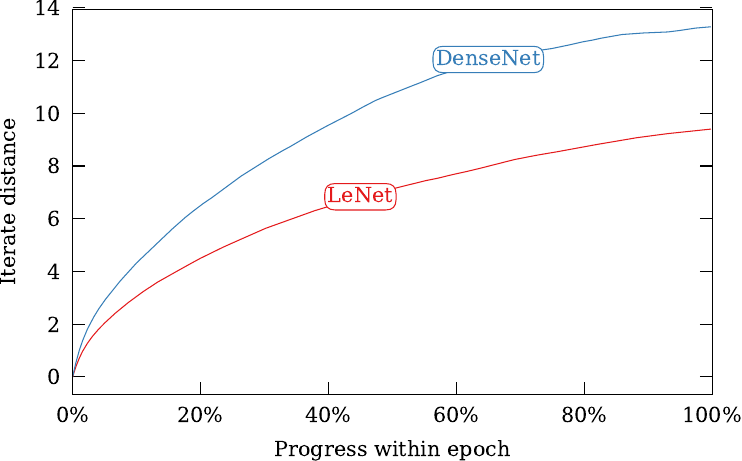}\includegraphics[width=0.48\columnwidth]{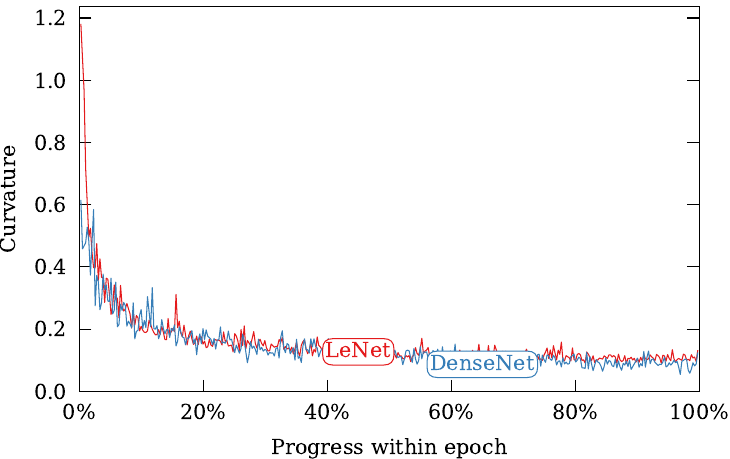}\vspace{-0.5em}
\par\end{centering}
\caption{\label{fig:iterate-distance}Distance moved from the snapshot point,
and curvature relative to the snapshot point, at epoch 50.}
\end{figure*}

Figure \ref{fig:iterate-distance} sheds further light on this. The
left plot shows how rapidly the current iterate moves within the same
epoch for LeNet and DenseNet models when training using SVRG. The
distance moved from the snapshot point increases significantly faster
for the DenseNet model compared to the LeNet model.

In contrast the right plot shows the curvature change during an epoch,
which we estimated as:\vspace{-0.5em}
\[
{\textstyle \left\Vert \frac{1}{\left|S_{i}\right|}\sum_{j\in S_{i}}\left[f_{j}^{\prime}(w_{k})-f_{j}^{\prime}(\tilde{w})\right]\right\Vert /\bigl\Vert w_{k}-\tilde{w}\bigr\Vert,}
\]
where $S_{i}$ is a sampled mini-batch. This can be seen as an empirical
measure of the Lipschitz smoothness constant. Surprisingly, the measured
curvature is very similar for the two models, which supports the idea
that iterate distance is the dominating factor in the lack of variance
reduction. The curvature is highest at the beginning of an epoch because
of the lack of smoothness of the objective (the Lipschitz smoothness
is potentially unbounded for non-smooth functions).

Several papers have show encouraging results when using SVRG variants
on small MNIST training problems \citep{svrg,nonconvex-scsg}. Our
failure to show any improvement when using SVRG on larger problems
should not be seen as a refutation of their results. Instead, we believe
it shows a fundamental problem with MNIST as a baseline for optimization
comparisons. Particularly with small neural network architectures,
it is not representative of harder deep learning training problems.\vspace{-0.8em}

\subsection{Smoothness\vspace{-0.7em}
}

Since known theoretical results for SVRG apply only to smooth objectives,
we also computed the variance when using the ELU activation function
\citep{elu}, a popular smooth activation that can be used as a drop-in
replacement for ReLU. We did see a small improvement in the degree
of variance reduction when using the ELU. There was still no significant
variance reduction on the DenseNet model.\vspace{-0.5em}

\section{Streaming SVRG Variants\vspace{-0.5em}
}

\label{sec:streaming-svrg}In Section \ref{sec:measuring}, we saw
that the amount of variance reduction quickly diminished as the optimization
procedure moved away from the snapshot point. One potential fix is
to perform snapshots at finer intervals. To avoid incurring the cost
of a full gradient evaluation at each snapshot, the class of streaming
SVRG \citep{streaming-svrg,nonconvex-scsg} methods instead use a
\emph{mega-batch} to compute the snapshot point. A mega-batch is typically
10-32 times larger than a regular mini-batch. To be precise, let the
mini-batch size be $b$ be and the mega-batch size be $B$. Streaming
SVRG alternates between computing a snapshot mega-batch gradient $\tilde{g}$
at $\tilde{w}=w_{k}$, and taking a sequence of SVRG inner loop steps
where a mini-batch $S_{k}$ is sampled, then a step is taken:
\begin{equation}
w_{k+1}=w_{k}-\gamma\left[\frac{1}{|S_{k}|}\sum_{i\in S_{k}}\left(f_{i}^{\prime}(w_{k})-f_{i}^{\prime}(\tilde{w})\right)+\tilde{g}\right].\label{eq:streaming-svrg}
\end{equation}
Although the theory suggests taking a random number of these steps,
often a fixed $m$ steps is used in practice, and we follow this procedure
as well.

In this formulation the data-points from the mega-batch and subsequent
$m$ steps are independent. Some further variance reduction is potentially
possible by sampling the mini-batches for the inner step from the
mega-batch, but at the cost of some bias. This approach has been explored
as the Stochastically Controlled Stochastic Gradient (SCSG) method
\citep{scsg}.

\begin{wrapfigure}{r}{0.5\columnwidth}%
\begin{centering}
\includegraphics[width=0.48\columnwidth]{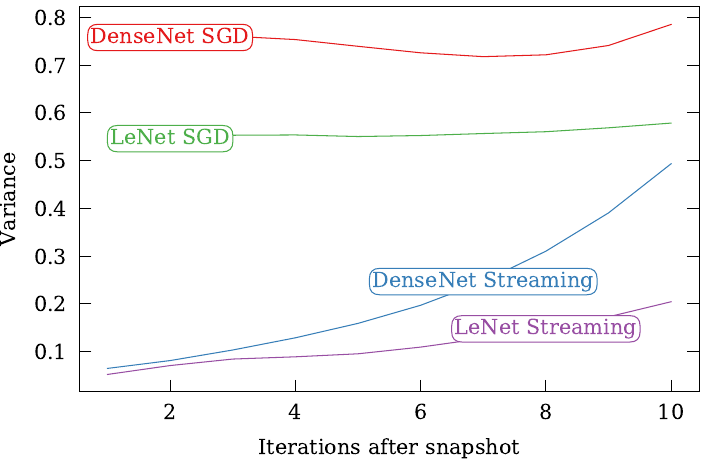}
\par\end{centering}
\caption{\label{fig:streaming-svrg-variance}Streaming SVRG Variance at epoch
50\vspace{-0.8em}
}
\end{wrapfigure}%
To investigate the effectiveness of streaming SVRG methods we produced
variance-over-time plots. We look at the variance of each individual
step after the computation of a mega-batch, where our mega-batches
were taken as 10x larger than our mini-batch size of 128 CIFAR10 instances,
and 10 inner steps were taken per snapshot. The data augmentation
and batch norm reset techniques from Section \ref{sec:complications}
were used to get the lowest variance possible. The variance is estimated
using the full dataset at each point. 

Figure \ref{fig:streaming-svrg-variance} shows the results at the
beginning of the 50th epoch. In both cases the variance is reduced
by 10x for the first step, as the two mini-batch terms cancel in Equation
\ref{eq:streaming-svrg}, resulting in just the mega-batch being used.
The variance quickly rises thereafter. These results are similar to
the non-streaming SVRG method, as we see that much greater variance
reduction is possible for LeNet. Recall that the amortized cost of
each step is three times that of SGD, so for the DenseNet model the
amount of variance reduction is not compelling.

\section{Other methods for non-convex variance-reduced optimization}

Although dozens of methods based upon the non-streaming variance reduction
framework have been developed, they can generally be characterized
into one of several classes: SAGA-like \citep{adefazio-nips2014},
SVRG-like \citep{svrg}, Dual \citep{SDCA}, Catalyst \citep{catalyst}
or SARAH-like \citep{sarah}. Each of these classes has the same issues
as those described for the basic SVRG, with some additional subtleties.
SAGA-like methods have lower computational costs than SVRG, but they
similar convergence rates on a per-epoch basis both empirically and
theoretically. As we show in the next Section, even on a per-epoch
basis and ignoring additional costs, SVRG doesn't improve over SGD
for large models, so we would not expect SAGA to show improvement
either. On such large models, SAGA is also impractical due to its
gradient storage requirements. 

Dual methods require the storage of dual iterates, resulting in similar
storage costs to SAGA, and are not generally applicable in the non-convex
setting. Most accelerated methods for the convex case fall within
the dual setup. 

Catalyst methods involve using a secondary variance-reduction method
to solve subproblems, which provides acceleration in the convex case.
Catalyst methods do not match the best theoretical rates in the general
non-convex case \citep{pmlr-v84-paquette18a}, and are not well-suited
to non-smooth models such as the ReLU-based neural networks used in
this work.

The SARAH approach is quite different from the other approaches described
above, but it suffers from the same high per-epoch computational cost
as SVRG that limits it's effectiveness, as it also uses two minibatch
evaluations each step together with a snapshot full gradient evaluation.
The SARAH++ variant \citep{sarahplusplus} has the best theoretical
convergence rate among the methods considered for non-convex problems.
However, we were not able to achieve reliable convergence with SARAH-style
methods on our test problems, which we attribute to an accumulation
of error in the inner loop.

\section{Convergence rate comparisons\vspace{-0.5em}
}

\label{sec:convergence-rate-comparisons}
\begin{figure*}
\begin{centering}
\subfloat[{\footnotesize{}LeNet on CIFAR10}]{\begin{centering}
\includegraphics[width=0.49\columnwidth]{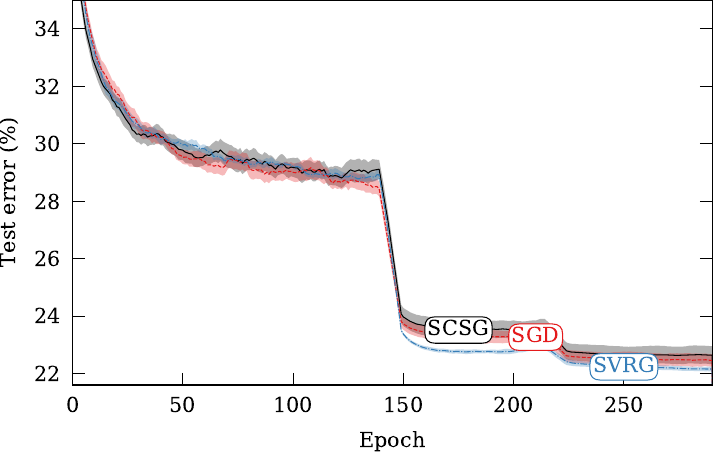}
\par\end{centering}
}\subfloat[{\footnotesize{}DenseNet on CIFAR10}]{\begin{centering}
\includegraphics[width=0.49\columnwidth]{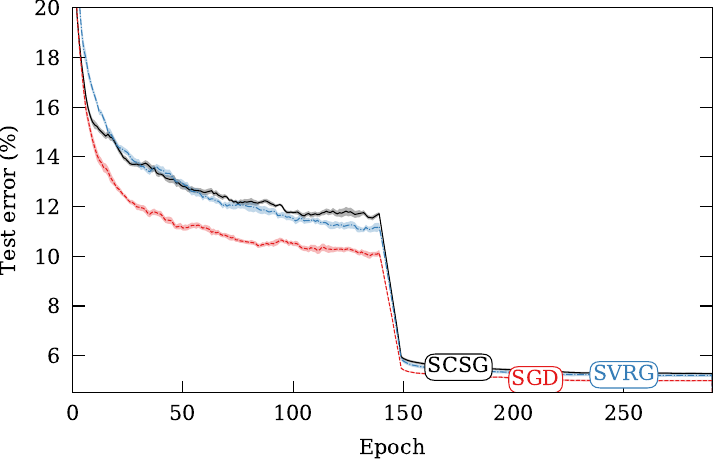}
\par\end{centering}
}
\par\end{centering}
\begin{centering}
\subfloat[{\footnotesize{}ResNet-110 on CIFAR10}]{\begin{centering}
\includegraphics[width=0.49\columnwidth]{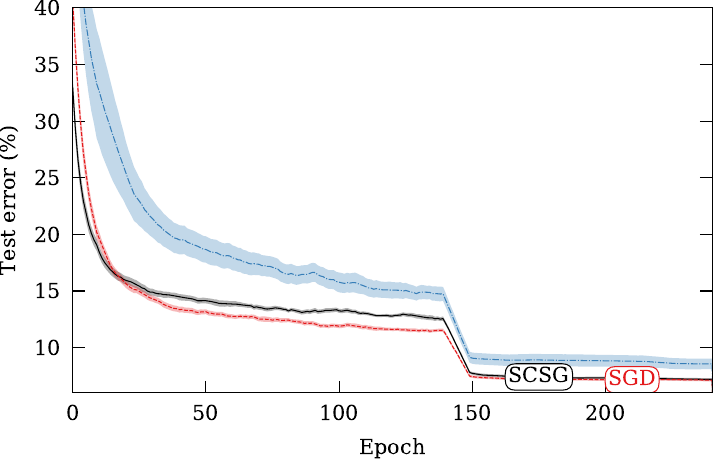}
\par\end{centering}
}\subfloat[\label{fig:Imagenet-comp}{\footnotesize{}ResNet-18 on ImageNet }]{\begin{centering}
\includegraphics[width=0.49\columnwidth]{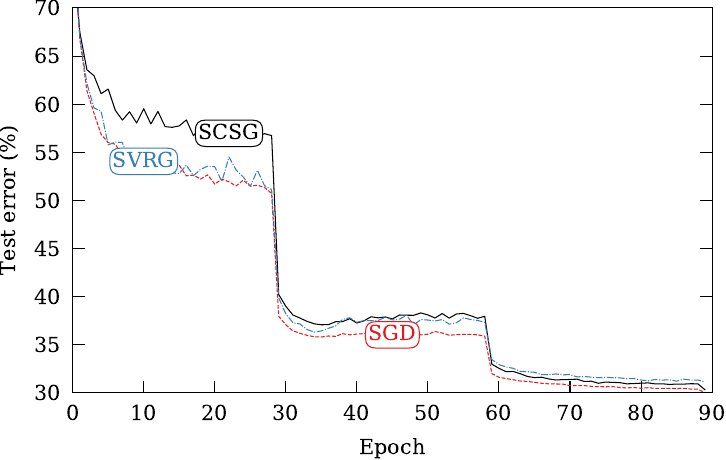}
\par\end{centering}
}
\par\end{centering}
\caption{\label{fig:error-comparison}Test error comparison between SGD, SVRG
and SCSG. For the CIFAR10 comparison a moving average (window size
10) of 10 runs is shown with 1 SE overlay, as results varied significantly
between runs. }
\end{figure*}
Together with the direct measures of variance reduction in Section
\ref{sec:measuring}, we also directly compared the convergence rate
of SGD, SVRG and the streaming method SCSG. The results are shown
in Figure \ref{fig:error-comparison}. For our CIFAR10 experiment,
an average of 10 runs is shown for each method, using the same momentum
(0.9) and learning rate (0.1) parameters for each, with a 10-fold
reduction in learning rate at epochs 150 and 225. We were not able
to see any improvement from using alternative hyper-parameters for
each method. A comparison was also performed on ImageNet using a ResNet-18
architecture and a single run for each method. Run-to-run variability
is much lower for image-net.

The variance reduction seen in SVRG comes at the cost of the introduction
of \emph{heavy correlation between consecutive steps}. This is why
the reduction in variance does not have the direct impact that increasing
batch size or decreasing learning rate has on the convergence rate,
and why convergence theory for VR methods requires careful proof techniques.
It is for this reason that the amount of variance reduction in Figure
\ref{fig:streaming-svrg-variance} doesn't necessarily manifest as
a direct improvement in convergence rate in practice. On the LeNet
problem we see that SVRG converges slightly faster than SGD, whereas
on the larger problems including ResNet on ImageNet (Figure \ref{fig:Imagenet-comp})
and DenseNet on CIFAR10 they are a little slower than SGD . This is
consistent with the differences in the amount of variance reduction
observed in the two cases in Figure \ref{fig:variance-ratios}, and
our hypothesis that SVRG performs worse for larger models. The SCSG
variant performs the worst in each comparison.\vspace{-0.8em}

\section{Fine-tuning with SVRG}

\begin{figure}
\begin{centering}
\subfloat[ResNet-50]{\centering{}\includegraphics[width=0.49\columnwidth]{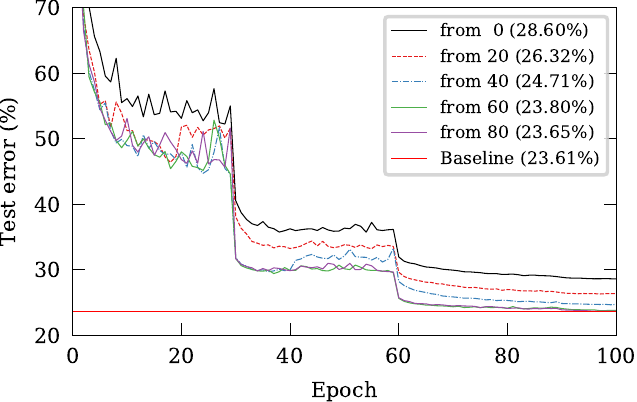}}\subfloat[DenseNet-169]{\begin{centering}
\includegraphics[width=0.49\columnwidth]{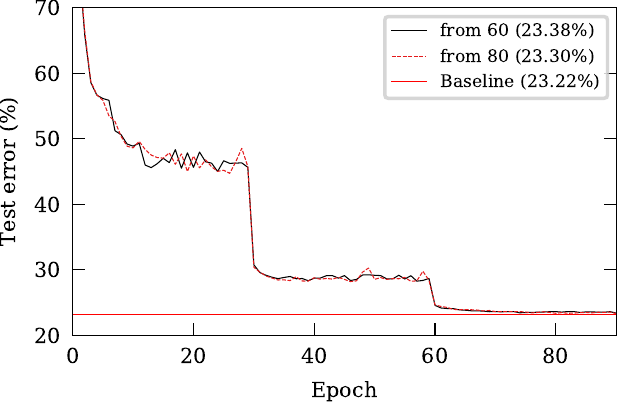}
\par\end{centering}

}\caption{\label{fig:fine-tuning}Fine-tuning on ImageNet with SVRG}
\par\end{centering}
\end{figure}
As we have shown that SVRG appears to only introduce a benefit late
in training, we performed experiments where we turned on SVRG after
a fixed number of epochs into training. Using the standard ResNet-50
architecture on ImageNet, we considered training using SVRG with momentum
from epoch 0, 20, 40, 60 or 80, with SGD with momentum used in prior
epochs. Figure \ref{fig:fine-tuning} shows that the fine-tuning process
did not lead to improved test accuracy at any interval compared to
the SGD only baseline. For further validation we evaluated a DenseNet-169
model, which we only fine-tuned from 60 and 80 epochs out to a total
of 90 epochs, due to the much slower model training. This model also
showed no improvement from the fine-tuning procedure.

\section*{Conclusion\vspace{-0.8em}
}

The negative results presented here are disheartening, however we
don't believe that they rule out the use of stochastic variance reduction
on deep learning problems. Rather, they suggest avenues for further
research. For instance, SVR can be applied adaptively; or on a meta
level to learning rates; or scaling matrices; and can potentially
be combined with methods like Adagrad \citep{adagrad} and ADAM \citet{adam}
to yield hybrid methods.

\bibliographystyle{plainnat}
\bibliography{vr}

\end{document}